\documentclass[letterpaper, 10 pt, conference]{ieeeconf}  

\usepackage[utf8]{inputenc}

\IEEEoverridecommandlockouts
\usepackage[american]{babel} 
\usepackage{cite}
\usepackage{float}
\usepackage{listings}
\usepackage{microtype}
\usepackage{booktabs}
\usepackage{url}
\usepackage{graphicx}
\usepackage{todonotes}
\usepackage{caption}
\usepackage{subcaption}
\usepackage{gensymb}

\usepackage{amssymb}
\usepackage{amsmath}
\usepackage{hyperref}
\usepackage{balance} 
\usepackage{float}

\title{\LARGE \bf
Rmagine: 3D Range Sensor Simulation in Polygonal Maps via Raytracing for Embedded Hardware on Mobile Robots
}

\author{Alexander Mock$^{1}$, Thomas Wiemann$^{2, 3}$ and Joachim Hertzberg$^{1, 3}$ 
\thanks{$^{1}$Knowledge Based Systems Group, Institute of Computer Science, Osnabrück University, Berghoffstraße 11, 49090 Osnabrück, Germany {\tt\small amock@uos.de}}%
\thanks{$^{2}$Autonomous Robotics Group, Institute of Computer Science, Osnabrück University, Berghoffstraße 11, 49090 Osnabrück, Germany {\tt\small twiemann@uos.de}}%
\thanks{$^{3}$DFKI Niedersachsen, Plan-based Robot Control, Berghoffstraße 11, 49090 Osnabrück, Germany {\tt\small thomas.wiemann@dfki.de, \tt\small joachim.hertzberg@dfki.de}}%
\thanks{The DFKI Niedersachsen (DFKI NI) is sponsored by the Ministry of Science and Culture of Lower Saxony and the VolkswagenStiftung}
}

\begin{document}

\maketitle
\thispagestyle{empty}
\pagestyle{empty}

\begin{abstract}

Sensor simulation has emerged as a promising and powerful technique to find solutions to many real-world robotic tasks like localization and pose tracking.
However, commonly used simulators have high hardware requirements and are therefore used mostly on high-end computers.
In this paper, we present an approach to simulate range sensors directly on embedded hardware of mobile robots that use triangle meshes as environment map. 
This library called Rmagine allows a robot to simulate sensor data for arbitrary range sensors directly on board via raytracing.
Since robots typically only have limited computational resources, the Rmagine aims at being flexible and lightweight, while scaling well even to large environment maps.
It runs on several platforms like Laptops or embedded computing boards like Nvidia Jetson by putting an unified API over the specific proprietary libraries provided by the hardware manufacturers. 
This work is designed to support the future development of robotic applications depending on simulation of range data that could previously not be computed in reasonable time on mobile systems.

\end{abstract}

\section{INTRODUCTION}

Triangle meshes are a standard data structure to represent 3D environments in many applications in computer graphics or robotics.
Today, it is possible to automatically generate triangle mesh maps even online on embedded systems~\cite{meisoldt21ecmr} that accurately capture the real world even in large scale environments~\cite{wiemann2018irc}.
This development allows to use such triangle meshes as an alternative map format to commonly used 2D grid maps or 3D voxel grids.
Recently, algorithms for mobile robot navigation were adapted to this format, where the triangle mesh was interpreted as 2D manifold graph to plan paths over the mesh surface to a given goal~\cite{puetz21cvp}.
Advances in the gaming industry now also make real time ray tracing feasible in such maps by using optimized algorithms and dedicated hardware.
This opens up novel applications besides graph-based approaches.
With the power of current raytracing techniques, it is possible to simulate sensor data even for complex range sensors like LiDARs in real time.
Any application that runs on a robot and at some point needs to answer the question "What would your sensor data look like if you were at that location?" could benefit from such 3D range sensor simulations.
For example, Monte Carlo localization (MCL) was recently adapted by Akai et al.~\cite{akai2020} to run in voxel maps by converting them to efficient distance fields.
Adapting MCL to triangle meshes could benefit from fast range sensor simulations during the importance sampling step.
Given a 6D localization in a triangle mesh, it would be also possible to separate static map parts from dynamic obstacles in the robots range measurements. 
Thus, it may be possible to implement novel change detection algorithms~\cite{rodrigues2020analytical} that work with triangles meshes as environmental maps.

\begin{figure}[t]
    \vspace{0.2cm}
    \centering
    \includegraphics[width=0.97\linewidth]{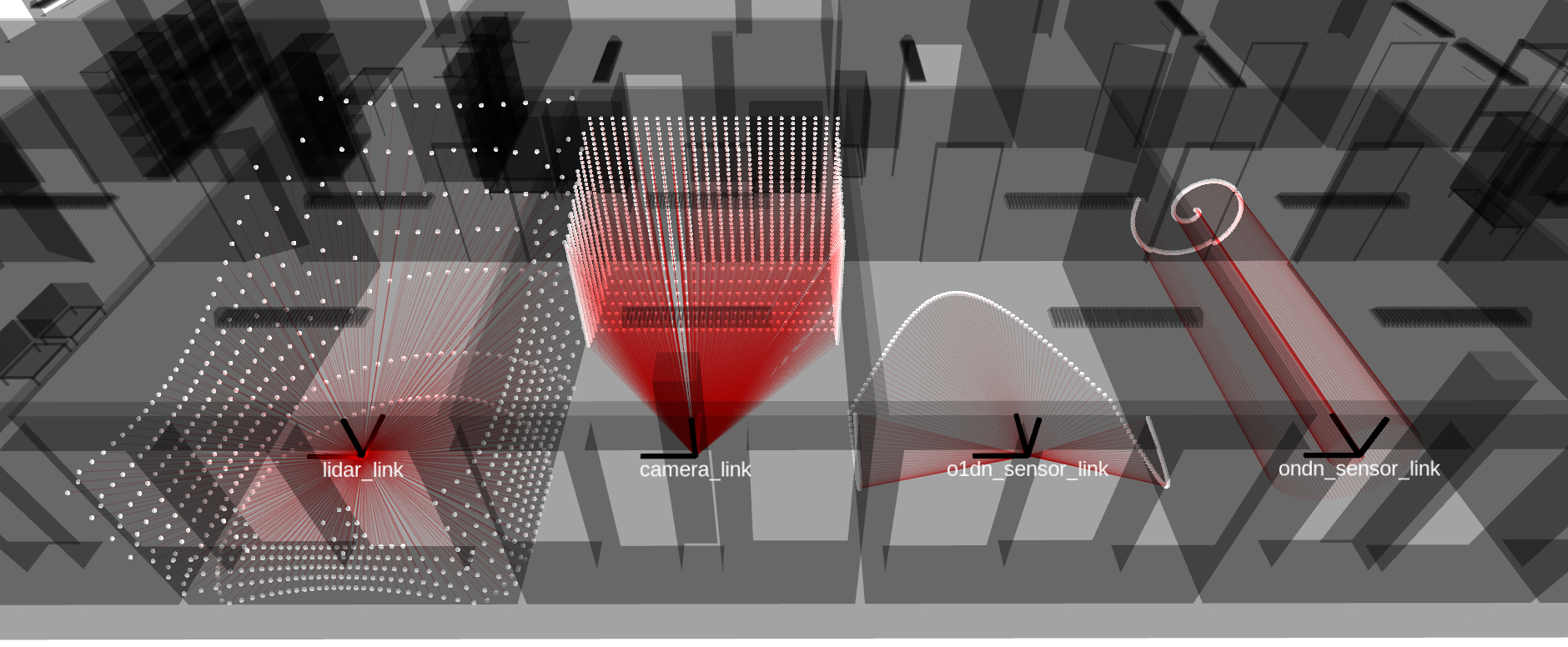}
    \includegraphics[width=0.97\linewidth]{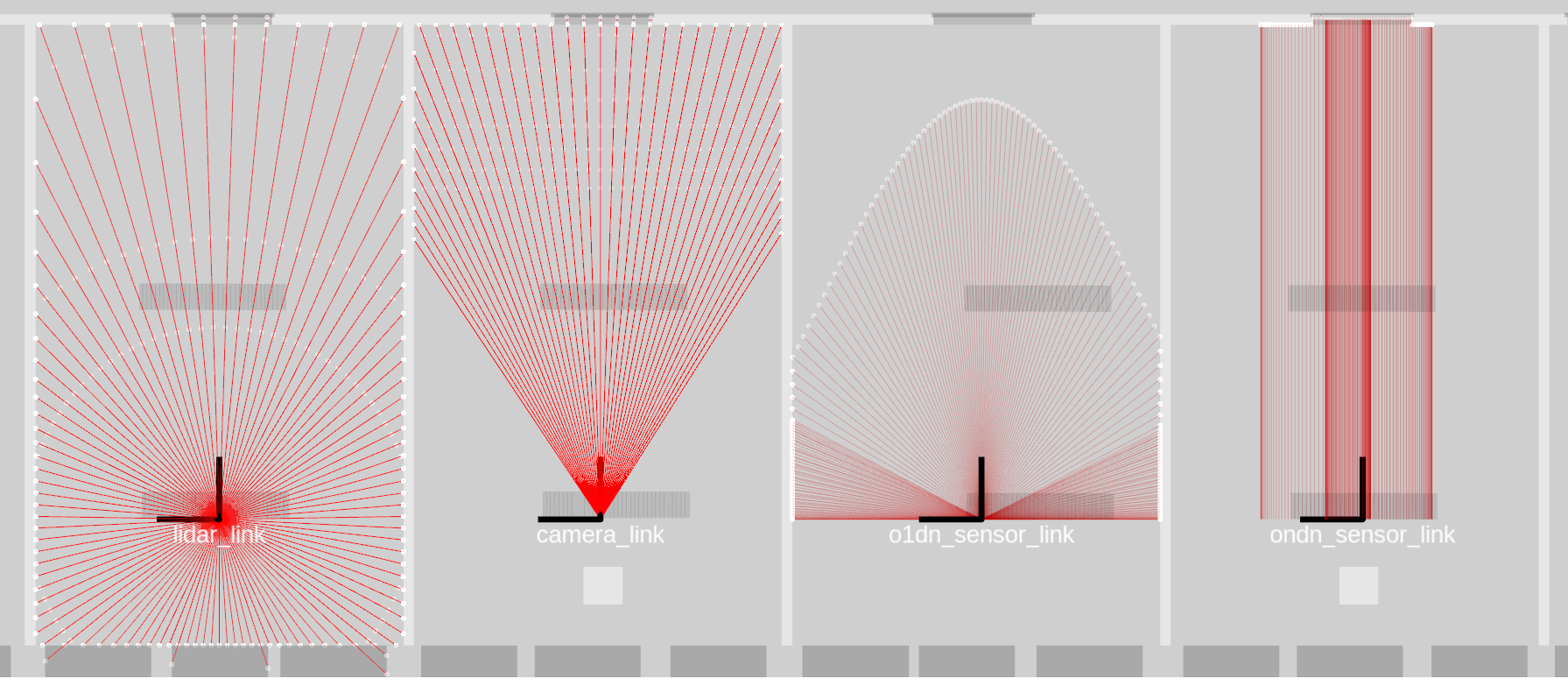}
    \caption{Rmagine simulations of several sensor models in a 3D map. The results are visualized with RViz. Top: 3D view, bottom: Top down 2D orthographic projection.}
    \label{fig:rvizsim}
    \vspace{-0.5cm}
\end{figure}

In this paper, we introduce our library called \emph{Rmagine}\footnote{Rmagine is released under BSD 3-Clause license and available at \url{https://github.com/uos/rmagine}.} that enables developers to simulate range sensors on the robot itself even in large-scale triangle meshes.
In contrast to commonly used simulators, Rmagine was designed to work as a standalone library without the requirement of a graphical user interface or rendering environment.
It focuses on a lightweight implementation that allows fast and parallel ray computations on different computational devices as CPUs and GPUs.
In contrast to classical simulators, it is not intended to simulate complex physics, but aims to provide efficient raytracing support on mobile device for future localization and tracking algorithms that rely on polygonal environment representations.
When hardware-acceleration is available by RTX GPUs, Rmagine is able to simulate thousands of scans from standard sensors per second.
Besides the hardware abstraction layer, Rmagine provides a number of built-in sensor models such as RGB-D cameras and LiDARs and an software interface that allows to implement any scan pattern as required by the application at hand.
Furthermore, we provide a ROS interface and Gazebo plugins\footnote{Rmagine Gazebo plugins is a software package released under BSD 3-Clause license and available at \url{https://github.com/uos/rmagine_gazebo_plugins}.} to showcase examples for this library in robotic contexts. 
Fig.~\ref{fig:rvizsim} shows the results of using Rmagine to simulate different range sensors in a 3D map rendered in RViz.
In our experiments, we demonstrate the benefits and limits of Rmagine and comparing it to the state-of-the art simulators ISAAC Sim~\cite{isaac2018} and Gazebo~\cite{koenig2004design}.

\section{RELATED WORK}




In computer graphics (CG), rendering is the process of generating the image of a virtual camera in a virtual scene.
The research field of classical photorealistic rendering mainly aims to render images that appear as realistic as possible to a human viewer. 
Contrary, simulations aim to be as physically correct as possible compared to real sensor data.
Physical correctness and realistic rendering however are often not so far apart.
The research field Physically Based Rendering (PBR), seeks to improve light-physics in order to let the renderings appear more realistic to human viewers~\cite{greenberg1997}.
Researchers of PBR early searched for a rendering strategy alternating to classical rasterization that models the transport of light more accurately in terms of physics.
The most promising approach is raytracing, which simulates light rays in a scene, taking into account reflections, transmissions, and emissions at the surface.
Due to the long computation time of the light traversals, for many years this technique has not been used in real-time applications.
In recent years algorithms accelerating scene traversals with BVH-trees on CPU and GPU improved significantly, especially when implemented in hardware on a GPU~\cite{keller2019we}.
Such realtime capable and physically correct raytracing triggered novel developments in photorealistic rendering and sensor simulation.
In 2018, NVIDIA released the Isaac SDK~\cite{isaac2018} that includes the simulator ISAAC Sim to simulate robots in 3D scenes.
ISAAC Sim utilizes NVIDIAs hardware accelerated raytracing to improve runtime and accuracy of physical simulations including sensor simulations. 
To simulate physics, it uses NVIDIAs PhysX API.
However, the software can only be used on computers with built-in NVIDIA RTX GPUs.
NVIDIA advertises ISAAC Sim for having such a high level of render realism that it could even be used for reinforcement learning using simulated data~\cite{makoviychuk2021isaac}.
An established alternative to ISAAC Sim is Gazebo, that has been successfully used in robotic applications~\cite{melo2019simcomp}.
Gazebo introduced the support of PBR with version 11, but the used OGRE rendering engine does not support raytracing.
To compensate for that, with Ignition-Gazebo they integrated the NVIDIA OptiX~\cite{optix10} raytracing engine in addition to OGRE.
This software is currently in an experimental state.
Besides ISAAC Sim and Gazebo there are many more approaches out there for general purpose sensor simulations.
LiDARSim \cite{manivasagam2020lidarsim}, for example, generates LiDAR simulations by raycasting in occupancy grids and enhances them with real sensor data using deep learning.

Applications running on robots such as localization \cite{akai2020}, change detection~\cite{rodrigues2020analytical}, or online inspection planning~\cite{abircher16ar} require very frequent spawns and removals of sensors at certain locations in the map.
Using ISAAC Sim or Gazebo for these problems, however, leads to large computational overheads.
In addition, ISAAC Sim and Gazebo must first be installed on a mobile robot, which may not be possible depending on the limitations of its computing units.
From the point of view of simple algorithmic raytracing operations, these simulators come with a significant overhead, as they also provide complex physical simulations and complex user interfaces, that are not required when running within a standalone algorithmic library.
Furthermore, these software packages do not provide a suitable API for standalone applications.
Hence, both ISAAC Sim and Gazebo are not intended for use on the robots directly~\cite{monteiro2019simulating, koenig2004design}.

Rmagine was especially developed to bridge this gap between external simulation environments and the need to perform fast sensor simulation on the robots for different tasks.
Since the available computational devices on robots differ significantly in hardware architecture, ranging from small embedded SoCs to standard laptops with or without dedicated GPUs, Rmagine provides a high level API that abstracts from the underlying hardware to provide raytracing support on many devices and architectures.

\section{RMAGINE}

\begin{figure}[t]
    \centering
    \includegraphics[width=\linewidth]{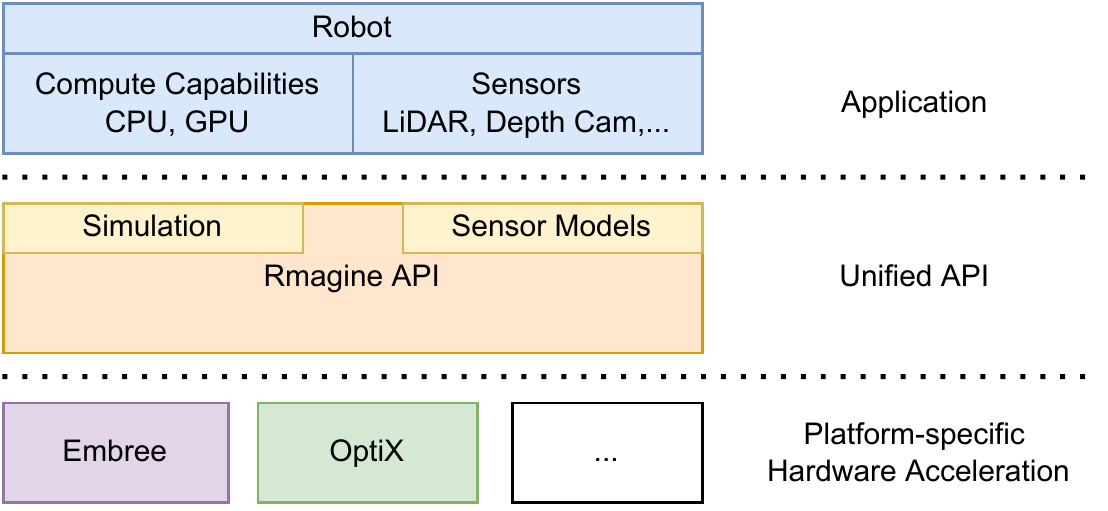}
    \caption{Rmagine API design. 
    A robot can choose which sensors to compute on which of its computing devices by selecting one of Rmagine's backends.}
    \label{fig:rmaginedesign}
\end{figure}

Rmagine is written in C++ and implements functions for simulating Range Sensors, accelerated with BVH trees, which are known to scale logarithmically in terms of render time with increasing map size\cite{gunther07}.
In addition, raycasting allows to model sensors more flexible and precise compared to traditional rasterization~\cite{gunther2019advantages}.
Currently, Rmagine uses Intel Embree~\cite{embree14} and NVIDIA OptiX\cite{optix10} as backends to support robots with CPUs and NVidia-GPUs equally.
Embree was improved for Intel CPUs, but can also be executed on other architectures.
To actually simulate data of a sensor, a map, a sensor model, and its pose relative to robot's base need to be provided.
Each of these elements can be placed and modified either in the CPU's RAM or in the GPU's VRAM. 
Consequently, custom implementations can update the map the sensor models and placements continuously either via CPU procedure calls or CUDA kernels during runtime.
The API of Rmagine is visualized in Fig. \ref{fig:rmaginedesign}.
The following examples can be tested with our Rmagine Gazebo plugins.

\subsection{Map}


In Rmagine, a map consists of a scene graph.
One scene holds many geometries or instances.
A geometry is, for example, a triangle mesh.
Each instance instantiates another subscene which again holds many geometries or instances.
Many instances can refer to a single subscene.
Instances of the same subscene may differ by their pose and scale in the scene.
It is possible to spawn many instances of the same subscene without actually duplicating their memory.
Rmagine implements one scene graph for each backend.
The backends provide similar interfaces differ internally in storing the scene data.
An Embree scene holds its elements in RAM while the OptiX scene stores the geometry data as CUDA buffers.
The only interface difference is, that the Optix scene can hold either instances or Geometries while the Embree scene can handle instances and geometries in the same scene.

These scenes can either be constructed programmatically or loaded from file formats for triangle meshes or scene graphs in general.
In order to provide a unified interface that supports a wide range of file formats, we use the Open Assets Importer Library (Assimp)~\cite{schulze2012open} to load meshes and scene graphs into our software.
The Assimp scene graph is then internally converted into Embree or OptiX scenes.
The respective acceleration structures are then constructed over the Rmagine scene graph.


Once a map is constructed, it also can be modified during runtime.
Modifications are, for example, adding, removing, rotating, moving, scaling instances or geometries.
Geometry modifications, such as moving single vertices, are also possible during runtime.
OptiX scene elements can also be changed using CUDA kernels, which allows a developer to animate scenes without any overheads in copy operations.
After committing the modifications of the map, Rmagine updates the acceleration structures.
Mostly, they do not require complete rebuilds and are updated partially at the tree depths effected by the committed modifications.
All these functionalities make the map dynamic, which allows to detect environmental changes and transfer them into the map.

\subsection{Sensor-Models}

\begin{table}[b]
    \centering
    \vspace{-0.3cm}
    \caption{Examples of well known sensors that can be modeled and simulated by Rmagine.}
    \label{tab:models}
    \begin{tabular}{ c c }
    \toprule
    Model & Sensors \\
    \midrule
    Spherical & Velodyne VLP-16, VLP-32, Ouster OS-0, OS-1, OS-2  \\ 
    Pinhole & MS Kinect, Asus Xtion, Intel RealSense D400 series \\ 
    O1Dn & Livox Mid-40, Avia  \\ 
    \bottomrule
    \end{tabular}
\end{table}

Robots may be equipped with several range sensors with different mathematical models. 
Rmagine implements the sensor models \texttt{Spherical}, \texttt{Pinhole}, \texttt{Cylindrical} and the custom models \texttt{O1Dn} and \texttt{OnDn}.
The \texttt{SphericalModel} consists of three values \texttt{theta}, \texttt{phi} and \texttt{range}, which represent polar angles $\theta$, $\varphi$ and the range $r$, respectively.
Since sensors that can be modeled with a spherical model typically have constant horizontal and vertical sampling increments, the values for $\theta$ and $\varphi$ can be discretized accordingly.
For example, a Velodyne VLP-16 LiDAR sensor has a $30\degree$ field of view with $16$ scan lines with a fixed resolution of $2\degree$.
Thus, $\varphi \in [-15.0\degree, +15.0\degree]$ occupies $16$ possible values with an increment of $2.0\degree$.
Horizontally, the VLP-16 has a $360\degree$ field of view with an adjustable resolution from $0.1\degree$ to $0.4\degree$.
Setting the horizontal resolution to $0.4\degree$, one scan line would provide $900$ single measurements.
Thus, $\theta \in [-360\degree, 360\degree)$ takes $900$ possible values with an increment of $0.4\degree$.
The measurements of a VLP-16 are valid up to a 100\,m distance: $r \in [0, 100]$.
Depth cameras can be modeled by a \texttt{PinholeModel} with the parameters image size $(w,h)$, focal lengths $(f_x, f_y)$ and principal point $(c_x, c_y)$.
In the last two models provided in Rmagine, \texttt{O1Dn} and \texttt{OnDn}, the origins and directions of the rays can be fully customized, allowing developers to implement sensors following arbitrary scan patterns. 
\texttt{O1Dn} has one origin and $N$ directions while \texttt{OnDn} has $N$ origins and $N$ directions where the $i$-th ray is constructed from the $i$-th origin and the $i$-th direction.
Tab.~\ref{tab:models} shows common sensor models that can be simulated with Rmagine.
Each simulated sensor can be attached to the virtual body of a robot by defining the transformation of the sensor to the robot's base $T_{sb}$.
Fig.~\ref{fig:rvizsim} visualizes simulations for each of Rmagine's sensor models.

\subsection{Simulation}

\begin{table}[b]
    \centering
    \vspace{-0.3cm}
    \caption{Requestable Attributes at map intersection.}
    \label{tab:intattr}
    \begin{minipage}{0.45\linewidth}
    \begin{tabular}{ c c c }
        \toprule
        Attr. & Type & Stride \\
        \midrule
        Hits & \texttt{uint8} & 1  \\ 
        Ranges & \texttt{float} & 1 \\ 
        Points & \texttt{float} & 3  \\
        Normals & \texttt{float} & 3 \\
        \bottomrule 
    \end{tabular}
    \end{minipage}
    \begin{minipage}{0.45\linewidth}
        \begin{tabular}{ c c c }
            \toprule
            Attr. & Type & Stride \\
            \midrule
            PrimIds & \texttt{uint32} & 1 \\ 
            GeomIds & \texttt{uint32} & 1 \\ 
            InstIds & \texttt{uint32} & 1 \\
            & &  \\
            \bottomrule 
        \end{tabular}
    \end{minipage}
    \vspace{-0.3cm}
\end{table}

\begin{figure}[b] 
    \centering
    \includegraphics[trim=0 70 0 200,clip,width=0.9\linewidth]{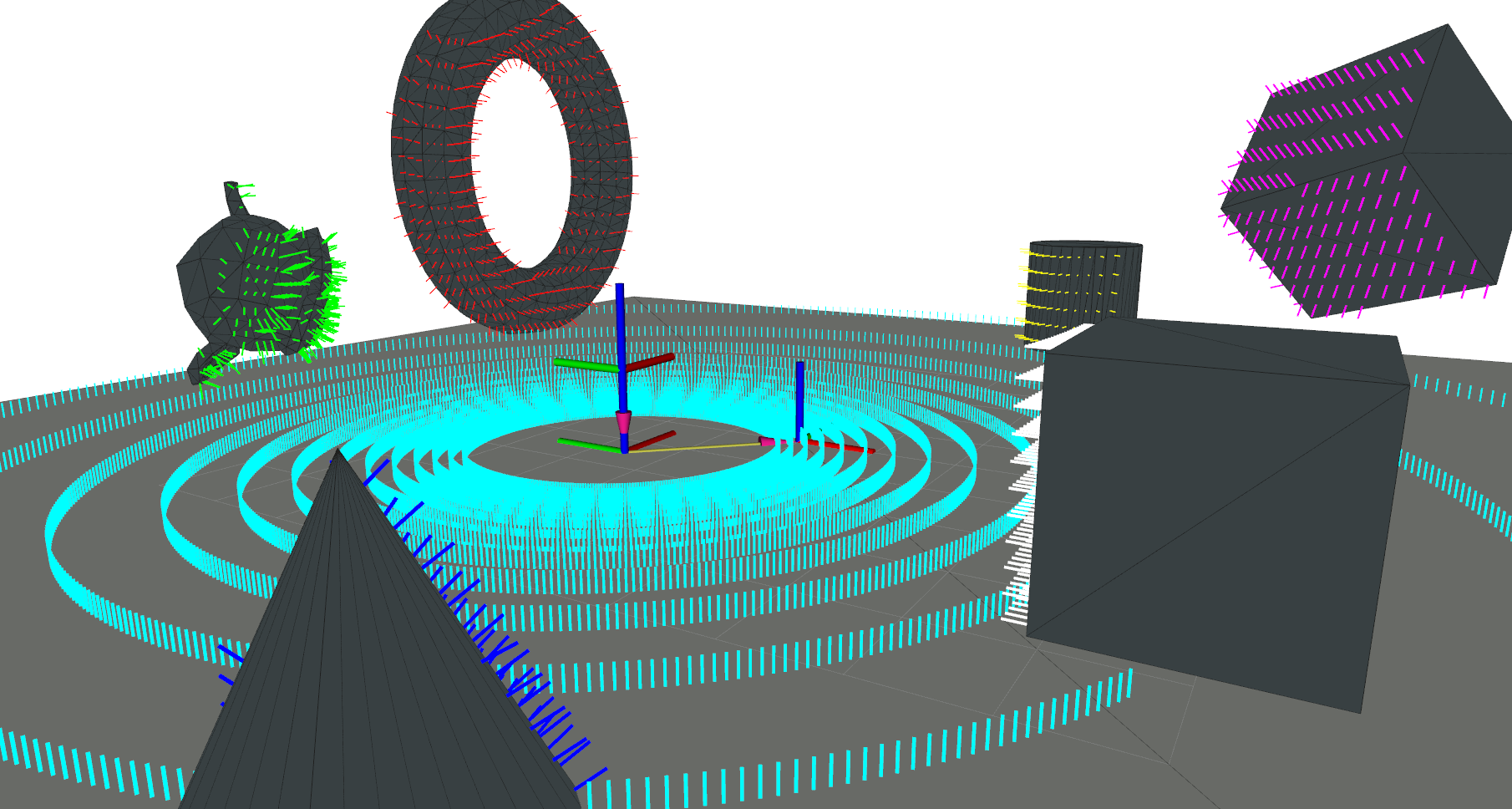}
    \caption{Rmagine simulation of a LiDAR in a 3D map consisting of several instances. Here, the intersection attributes Points, Normals and InstIds (colors) are queried.}
    \label{fig:attrviz}
    \vspace{-0.5cm}
\end{figure}

\begin{figure}[t]
    \centering
    \begin{subfigure}[t]{0.49\linewidth}
        \centering
        \includegraphics[trim=50 120 50 120,clip,width=0.99\linewidth]{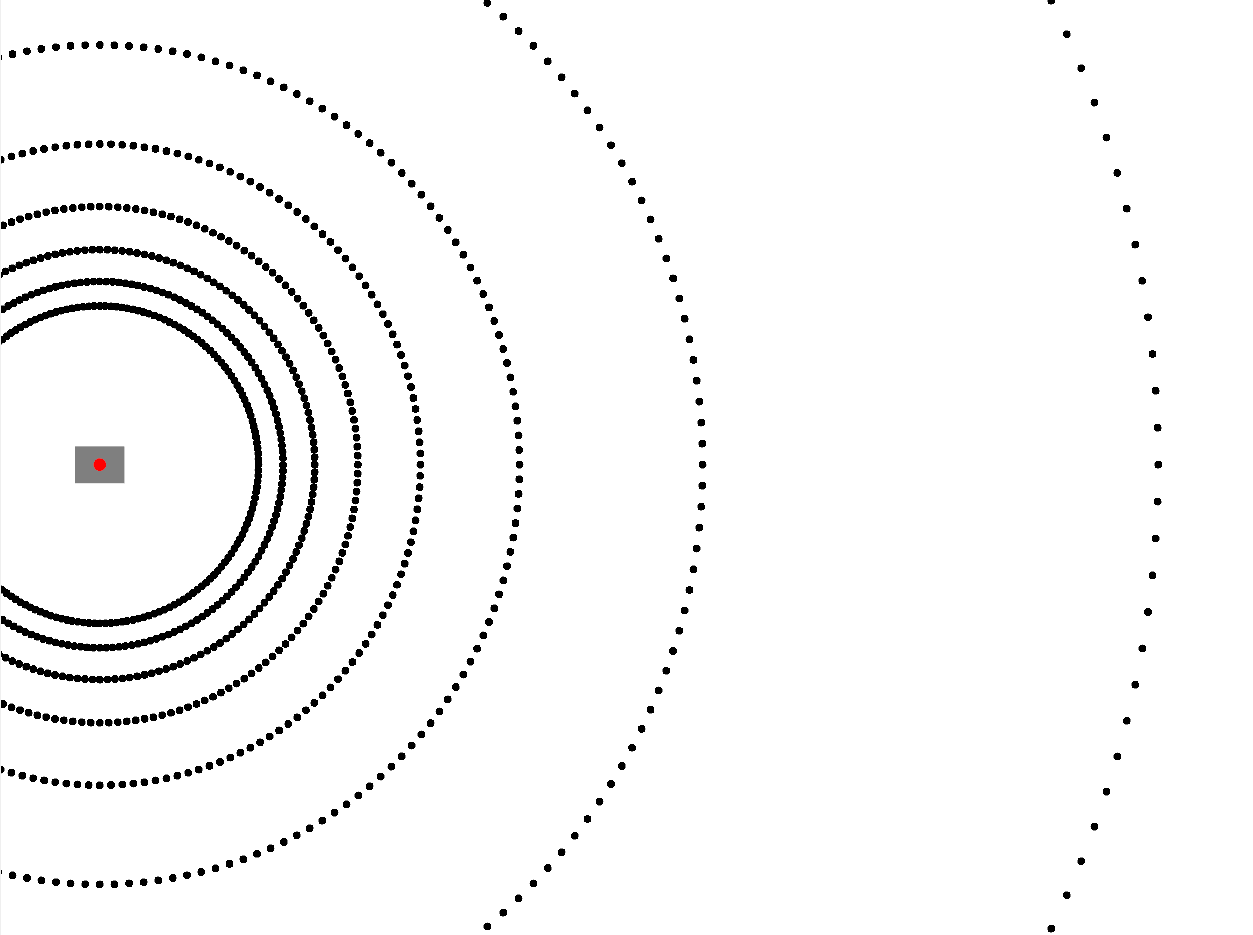} 
        \caption{No noise}
        \label{fig:noise:none}
    \end{subfigure}
    \begin{subfigure}[t]{0.49\linewidth}
        \centering
        \includegraphics[trim=50 120 50 120,clip,width=0.99\linewidth]{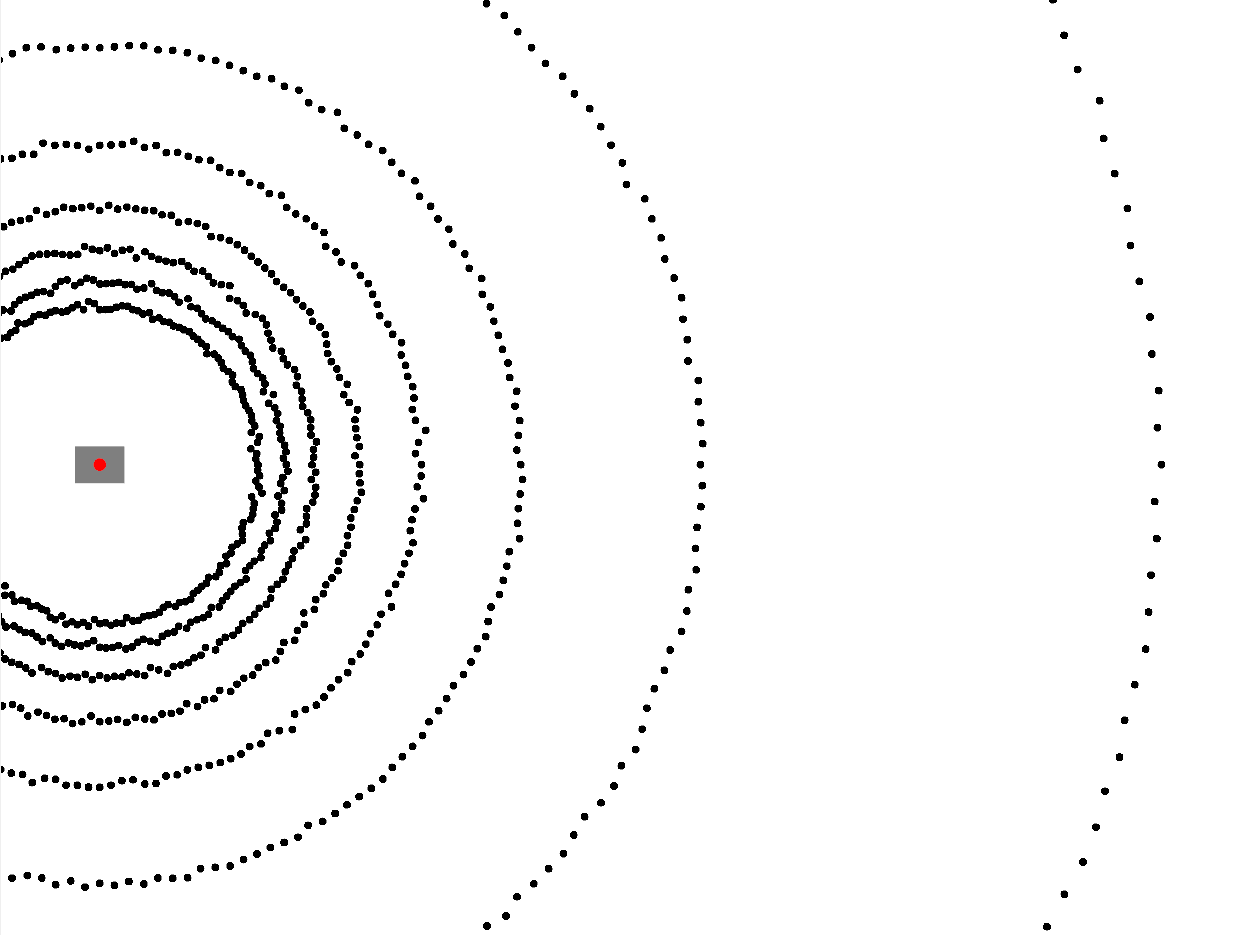} 
        \caption{Gaussian}
        \label{fig:noise:gaussian}
    \end{subfigure} \\
    \begin{subfigure}[t]{0.49\linewidth}
        \centering
        \includegraphics[trim=50 120 50 120,clip,width=0.99\linewidth]{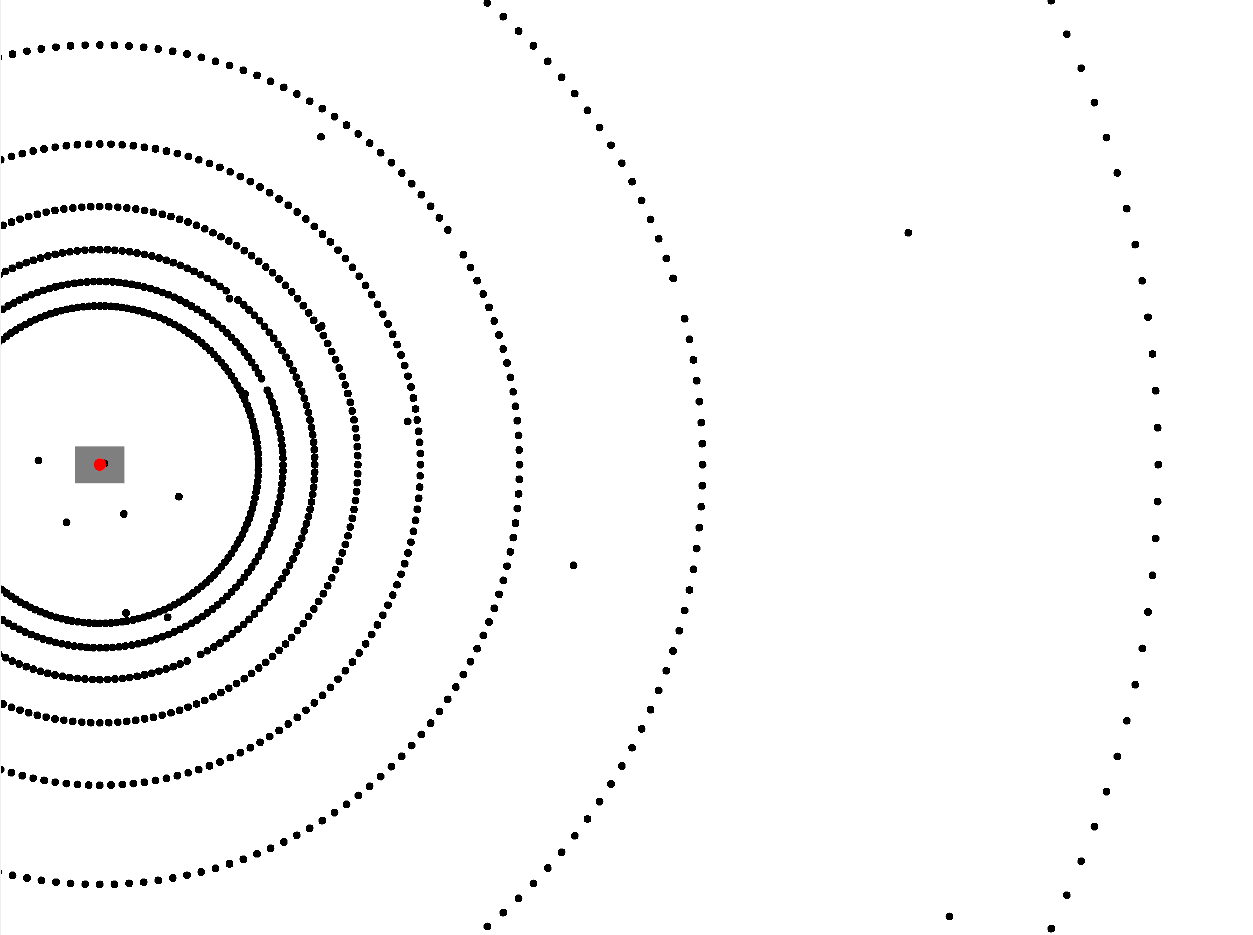}
        \caption{Dust}
        \label{fig:noise:dust}
    \end{subfigure}
    \begin{subfigure}[t]{0.49\linewidth}
        \centering
        \includegraphics[trim=50 120 50 120,clip,width=0.99\linewidth]{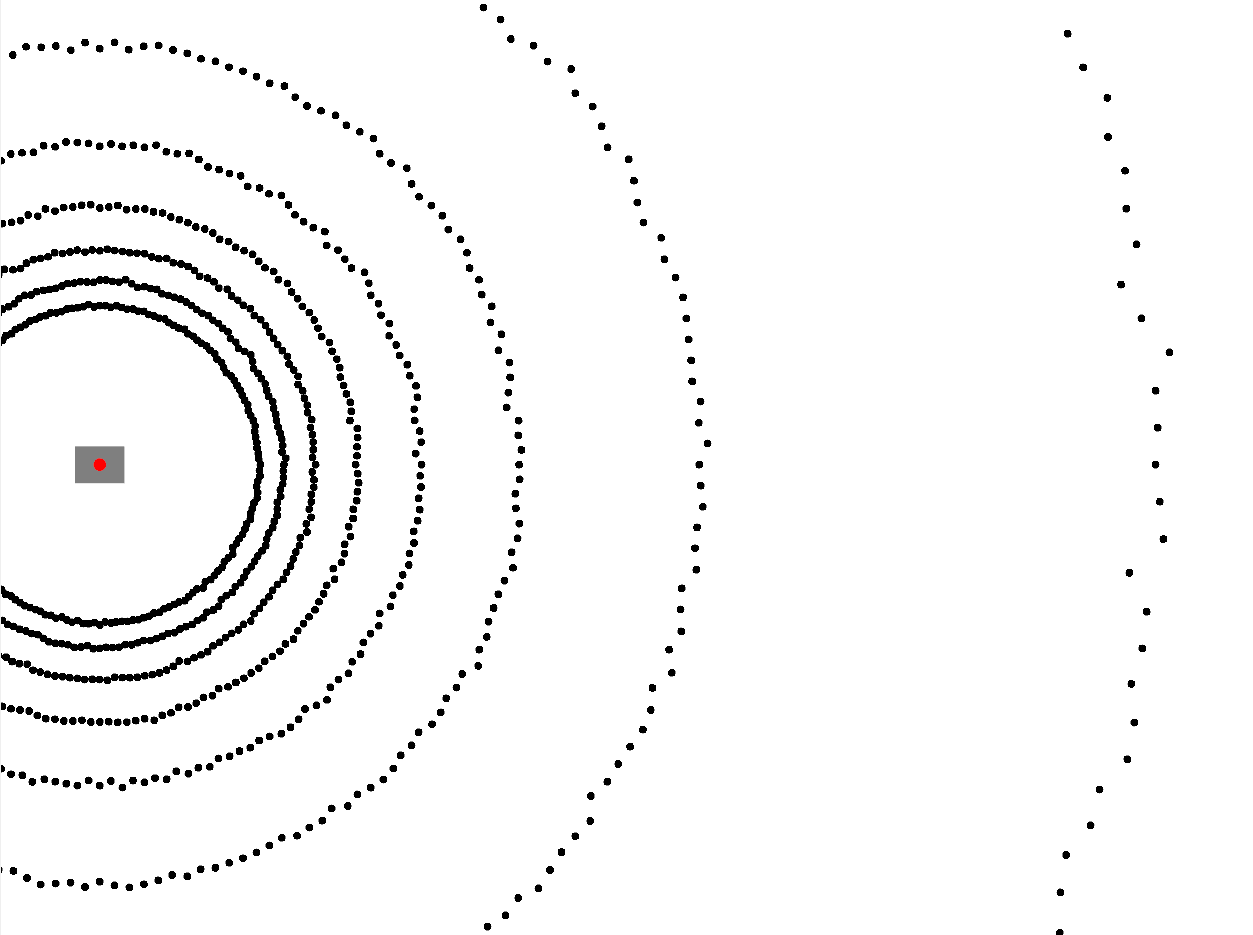}
        \caption{Gaussian (range dependent)}
        \label{fig:noise:relgaussian}
    \end{subfigure}
    \caption{Noise models provided by the Rmagine API.}
    \label{fig:noise}
    \vspace{-0.3cm}
        
\end{figure}

Given the Rmagine map and the sensors models, simulated data can be generated for arbitrary poses in the map.
First, a map is passed to the Sensor-Simulator module as well as the particular transformation from sensor to base \texttt{Tsb}.
This information together with the defined sensor model is used to simulate the intersections of the respective scan pattern with the 3D model.
Besides the Cartesian coordinates, further attributes of the intersection point can be requested if required.
The hit attribute can be either 1 or 0, depending on whether the ray intersects the map or not.
The range attribute represents the distance from the ray's origin to the first intersection with the map along the ray's direction.
Querying the normal attribute enables the calculation of the normal at the intersected face.
The ID of instance, geometry, or triangle can be retrieved by the InstId, GeomId, or FaceId attribute respectively.
A list of all available intersection attributes is summarized in Tab.~\ref{tab:intattr}.
The implementation automatically optimizes the OptiX and Embree code depending on which combination of intersection attributes where queried.
Thus, we assure the raycasting operation is performed only once per requested bundle of intersection attributes.
The resulting attribute buffers are located in the sensors reference frame.
Fig.~\ref{fig:attrviz} visualizes the simulation results querying the attributes Points, Normals and InstIds.
Depending on the device that computes the intersections, the results are stored either into buffer in CPU or GPU memory.
This allows other applications to flexibly post-process the simulated data without any overhead for transferring data between computing devices.
This way, Rmagine distributes the workloads flexibly to computing devices of the robot.
For example, Rmagine provides functions to apply several noise models to the simulated measurements. 
Rmagine selects the actual implementation of the function according to the memory location of the simulated measurements.
If the data is located in a CUDA buffer on the GPU, the random numbers are generated directly on the device by using cuRAND, CUDAs random number generation library. 
Otherwise, the functionalities of the C++ standard random number library are used.
Examples of noise models implemented in Rmagine are shown in Fig.~\ref{fig:noise}.

\subsection{Example Modeling}

\begin{figure}[t]
    \centering
    \includegraphics[width=0.9\linewidth]{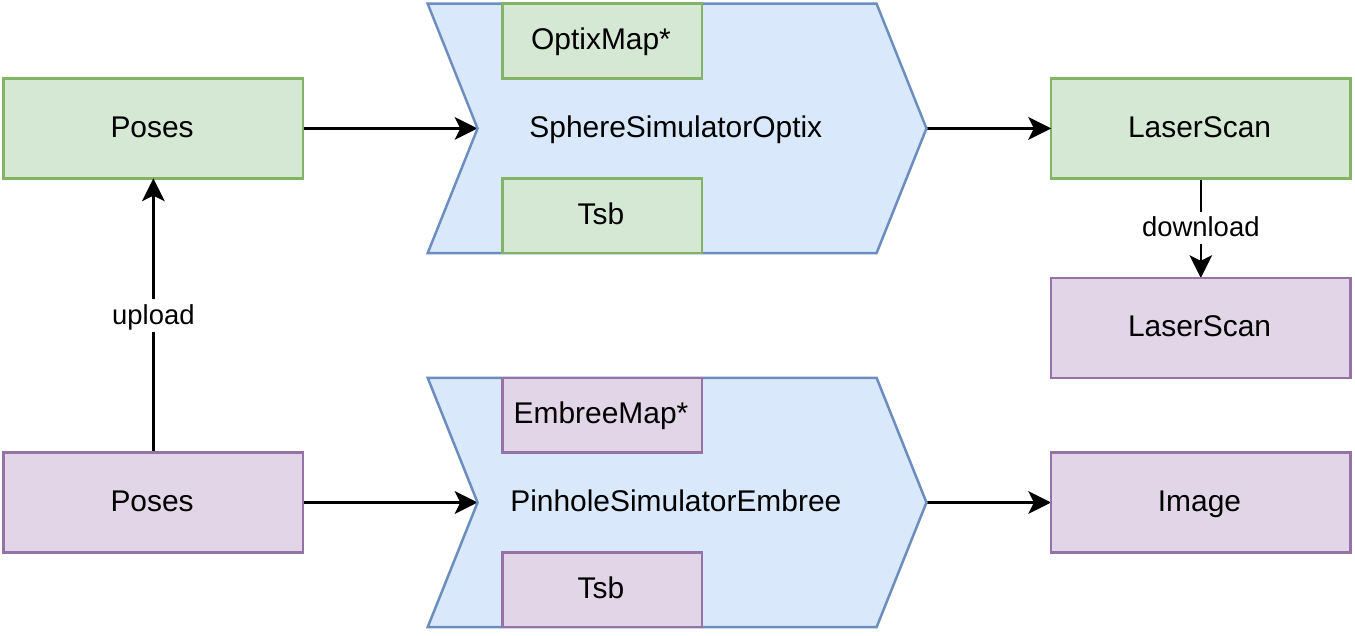}
    \caption{Internal structure of a Rmagine simulation of a robot equipped with one LiDAR and one depth camera. The depth camera is simulated on the CPU (purple) and the LiDAR on the GPU (green).}
    \label{fig:rmagine_example}
    \vspace{-0.3cm} 
\end{figure}

In this section, we present an example for modeling a LiDAR and a depth camera in Rmagine with GPU and CPU-based raytracing.
For this, we need to load the map data both to CPU and GPU.
The depth camera is modeled using suitable parameters for a pinhole model, the LiDARs is modeled as an instance of a spherical model.
The actual simulator classes are named by the sensor-model type X as prefix and the computing backend Y as suffix: \texttt{XSimulatorY}.
Accordingly, we need a \texttt{SphereSimulatorOptix} and a \texttt{PinholeSimulatorEmbree} to compute a LiDAR via OptiX and a depth camera via Embree.
For sensor data simulation, we define a set of possible poses that is loaded both to CPU and GPU memory.
After each simulation request, the scans simulated on the GPU are then downloaded into CPU memory.
The workflow of this example is sketched in Fig.~\ref{fig:rmagine_example}. 
Algorithms running on the GPU are shown in green, CPU computations are shown in purple.
As robots often execute several applications running in parallel, it might be desirable to support some mechanism for load balancing.
For that, Rmagine allows to switch the usage of computing devices during runtime.
Switching devices in this example can be realized by adding a \texttt{SphereSimulatorEmbree} and \texttt{PinholeSimulatorOptix} that can be called respectively.

\section{EXPERIMENTS}

This section evaluates Rmagine with respect to our design goals.
For this, we installed Rmagine on several test devices to prove the compatibility with different computing architectures and to compare it to Isaac Sim and Gazebo.
Furthermore, we present an in-depth runtime analysis of Rmagine in different environments.

\subsection{Compatibility}

Rmagine completely inherits the device compatibility of its raycasting backbones Intel Embree and NVIDIA OptiX.
Embree at version 3 runs on devices from Intel, AMD or on ARM CPUs. 
The developers state that it performs approximately 10\% better on a Intel CPU when compiled it with the Intel C++ compiler.
However, to allow a fair comparison, we compiled Embree with the GNU compiler (version 9.3.0) on an Ubuntu 20.04 operating system.
GPU support requires OptiX of version 7.2 or newer.
To support this in Rmagine, the OptiX headers need to be downloaded from NVIDIA, while the OptiX library itself is automatically installed through the NVIDIA drivers.
OptiX runs on both NVIDIA GTX and NVIDIA RTX graphics cards but not on GPUs of other manufacturers.
The OptiX host API can be installed on both Intel and AMD processors. 
Since version 7.3, OptiX also supports installation on ARM chips. 
For evaluation, we installed Rmagine on an office PC, Lenovo Legion 5 Pro 16ACH6H laptop, an Intel NUC RNUC11PHKi7CAA2, a Lenovo 20MF000XGE laptop and a NVIDIA Jetson Xavier NX developer board.
The specifications of the used devices are summarized in~Tab.~\ref{tab:devices}.

On the Jetson NX, however, we could not install the GPU support since the OptiX library is currently not deployed with the NVIDIA driver for Volta architectures.
Nevertheless, compared to the install specifications of ISAAC Sim and Gazebo, our software covers a wider range of compatible devices as shown in the left part of Tab.~\ref{tab:compinst}.
Regarding the required compute capabilities, (NV) stands for NVIDIA and (RT) for raytracing. 
A negation means non-NVIDIA or no raytracing support respectively.
In contrast to Rmagine, ISAAC Sim necessarily requires a dedicated GPU. 
However, Gazebo allows to simulate sensor data also on non-NVIDIA GPUs, which Rmagine currently does not support.
ISAAC Sim and Gazebo both implement interfaces to let developers receive the simulated sensor data on CPU side through buffers placed in RAM.
Rmagine additionally implements an interface to directly access the GPU simulations via CUDA buffers to avoid unnecessary downloads when passed to post-processing CUDA kernels afterwards.

\begin{table}[t]
    \centering
    \caption{Specifications of the devices that were used to evaluate Rmagine, i.e., a desktop PC (PC), a Lenovo Legion (Legion), a Intel NUC (NUC), a Lenovo X1 (X1) and a Jetson Xavier NX (NX).}
    \label{tab:devices}
    \begin{tabular}{ c c c }
    \toprule
    ID & CPU Type & NVIDIA GPU Type \\
    \midrule
    PC & Ryzen 7 3800X & RTX 2070 Super \\ 
    Legion & Ryzen 7 5800H & RTX 3070 Laptop \\ 
    NUC & i7 1165G7 & RTX 2060 Max-P \\ 
    X1 & i7 8750H & GTX 1050 Ti Max-Q \\ 
    NX & Carmel ARM 6C & Volta 48TC \\ 
    \bottomrule
    \end{tabular}
    \vspace{-0.3cm}
\end{table}




\subsection{Performance}

\begin{figure}[b]
    \centering
    \includegraphics[width=0.48\linewidth]{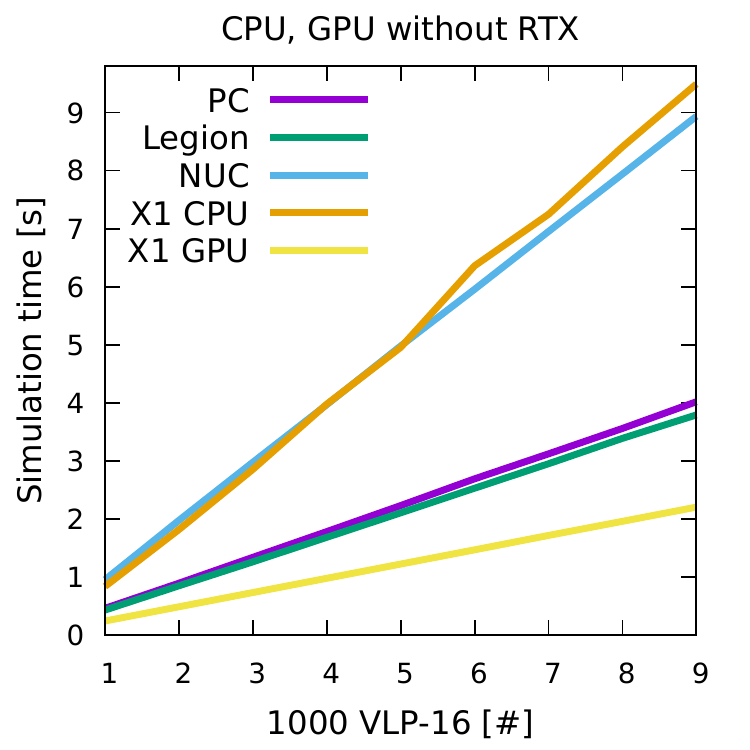}
    \includegraphics[width=0.48\linewidth]{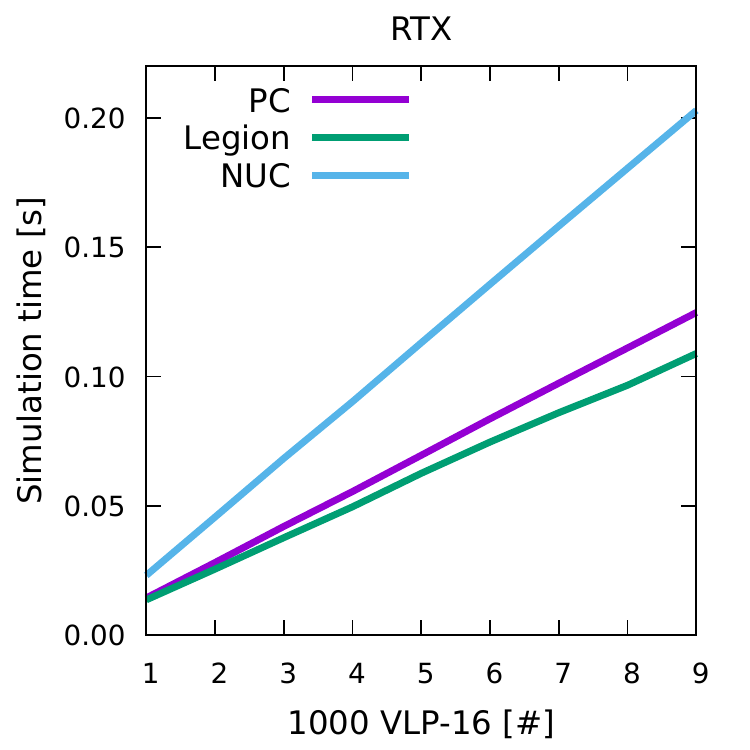} 
        
    \caption{Performance evaluation of Rmagine with a fixed map size of 1M triangles varying the number of VLP-16 scans, each consisting of 14,400 rays.
    The x-axis denotes $x \cdot 1,000$ number of scans that have been simulated by Rmagine.}
    \label{fig:runtime:sensors:rm}
\end{figure}


In the next experiments the performance of Rmagine was analyzed.
We evaluated Rmagine's runtime in detail using several devices and considering different computational loads and map sizes. 
In each of the following experiments, we simulated the Velodyne VLP-16 LiDAR sensor, which has 16 scan lines, with 900 range measurements each.
Per sensor frame, we thus needed to simulate 14,400 intersections with the map.
For the first evaluation we used a synthetic sphere with 1 million faces as map.
Inside this sphere, we placed multiple VLP-16 sensors at random poses.
The runtime measurements for increasing numbers of simulated poses are presented in Fig.~\ref{fig:runtime:sensors:rm}.
Each experiment was repeated 100 times with different distributions on all devices.
The results show that the runtime is linear to the number of simulated sensors.

\begin{figure*}[t] 
    \centering
    \begin{subfigure}{.49\linewidth}
        \centering
        \includegraphics[width=0.48\linewidth]{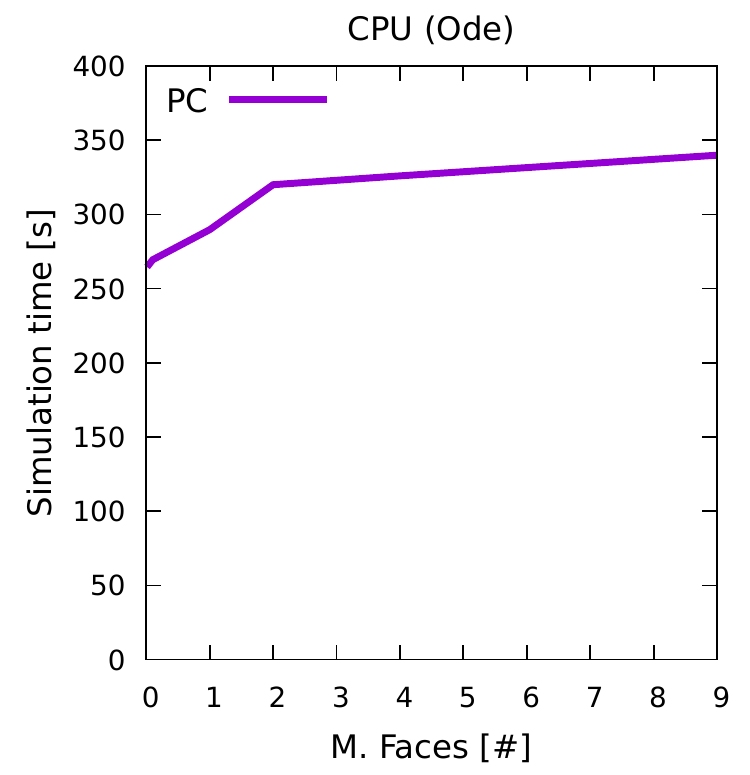}
        \includegraphics[width=0.48\linewidth]{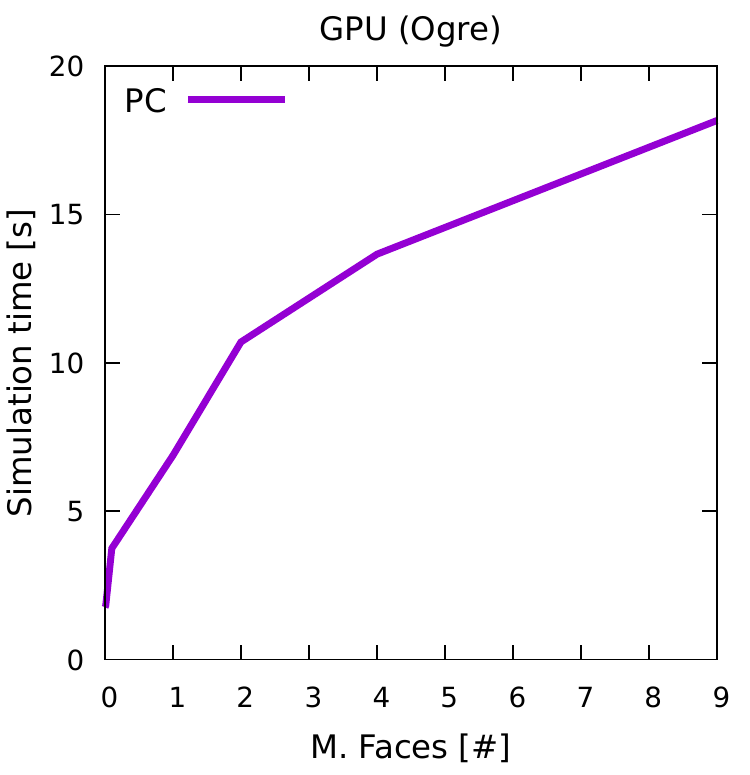}
        \vspace{-0.3cm}
        \caption{Gazebo}
        \label{fig:runtime:map:gazebo}
    \end{subfigure}
    \hfill
    \begin{subfigure}{.49\linewidth}
        \centering
        \includegraphics[width=0.48\linewidth]{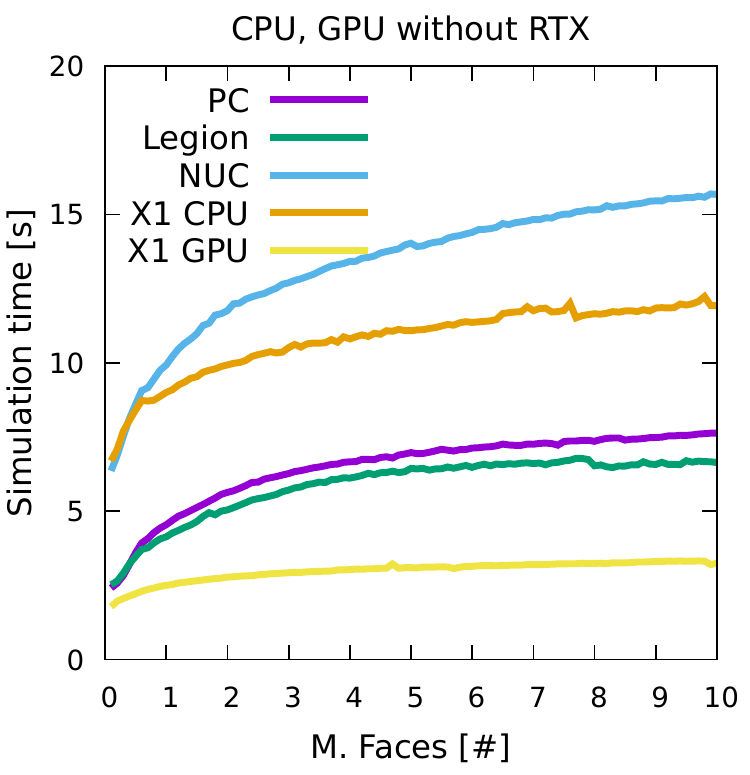}
        \includegraphics[width=0.48\linewidth]{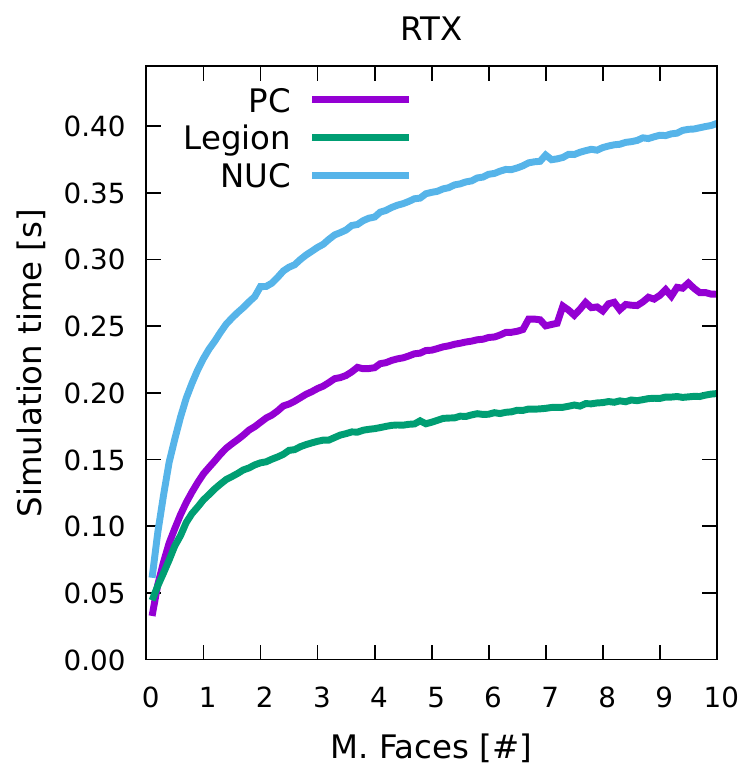}
        \vspace{-0.3cm}
        \caption{Rmagine}
        \label{fig:runtime:map:rmagine}
    \end{subfigure}
    \caption{Runtimes of simulating 10\,000 VLP-16 scans on different devices (Tab. \ref{tab:devices}) while varying the map's size (x-axis).
    The results of Rmagine are shown in (b).
    As baseline we measured the runtime for Gazebo as well (a).}
    \label{fig:runtime:map}
    \vspace{-0.2cm}
\end{figure*}

In the next experiment, we placed 10,000 VLP-16 sensors on static poses while varying the map size.
The entire experiment was again computed and repeated 100 times on our test devices.
The average runtimes are depicted in Fig.~\ref{fig:runtime:map}.
Since both Gazebo and Isaac Sim were originally not designed to run on mobile robot hardware, a performance comparison here would not be meaningful.
Therefore, we added a Gazebo baseline at Fig.~\ref{fig:runtime:map:gazebo} just for orientation.
Here, Rmagine outperforms Gazebo in every instantiation.
Since both Embree and OptiX internally use BVH-trees to accelerate ray traversals, we expect the simulation to scale well in larger maps.
Matching this expectation, the experimental results demonstrate the simulation time rises logarithmically with increasing map sizes.

\begin{figure}[t]
    \centering
    \begin{subfigure}[t]{0.49\linewidth}
        \includegraphics[trim=0 0 0 0,clip,width=\linewidth]{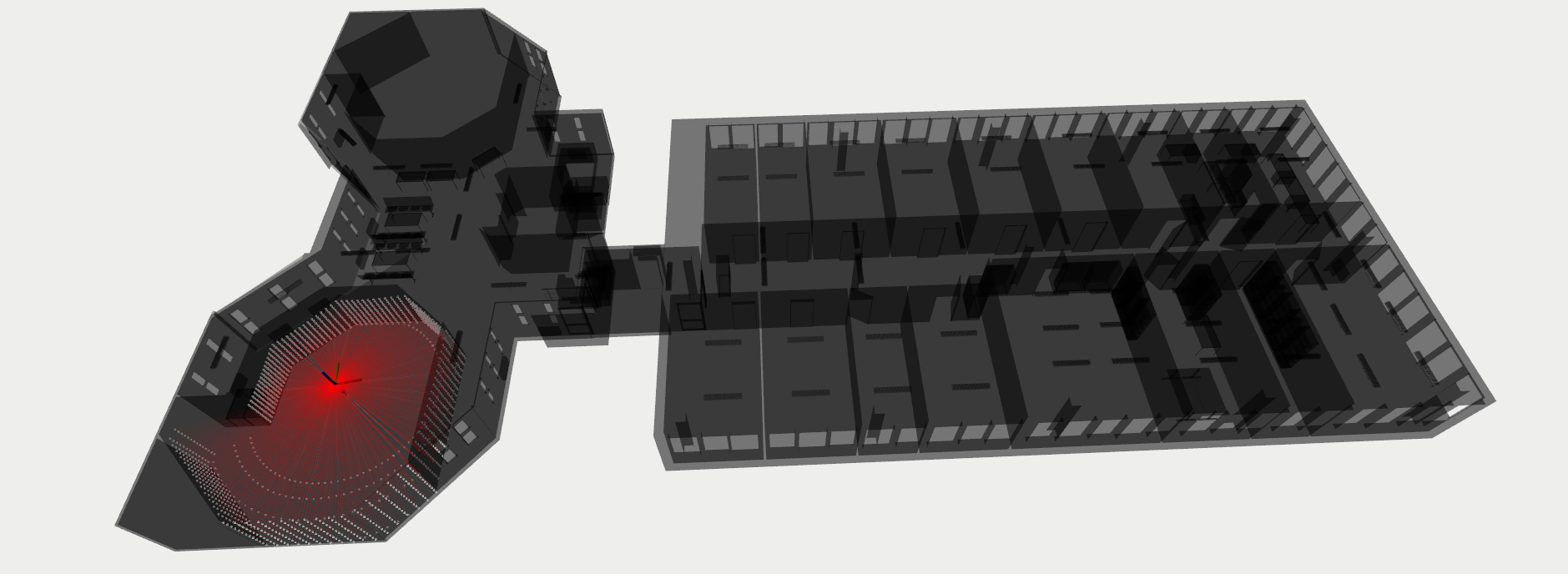}
        \caption{Indoor hand modeled}
        \label{fig:datasets:avz}
    \end{subfigure}
    \begin{subfigure}[t]{0.49\linewidth}
        \includegraphics[trim=0 0 0 0,clip,width=\linewidth]{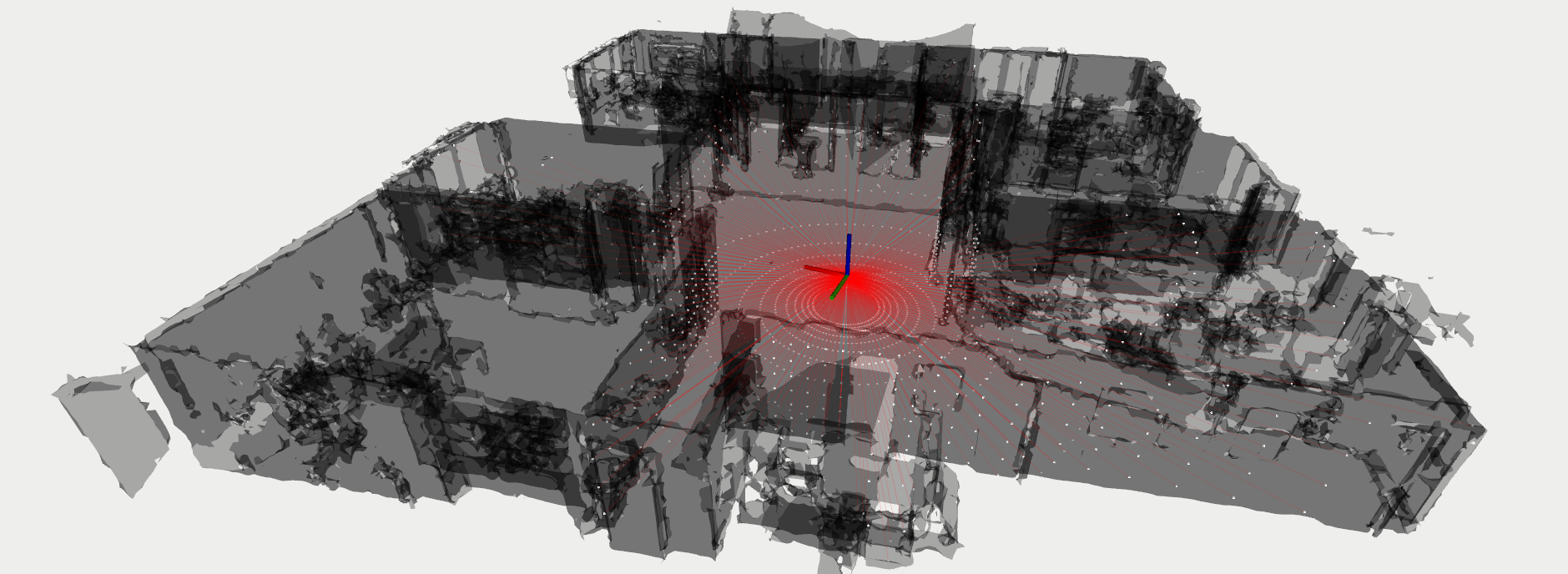}
        \caption{Indoor}
        \label{fig:datasets:berg}
    \end{subfigure} \\
    \begin{subfigure}[t]{0.49\linewidth}
        \includegraphics[trim=0 0 0 0,clip,width=\linewidth]{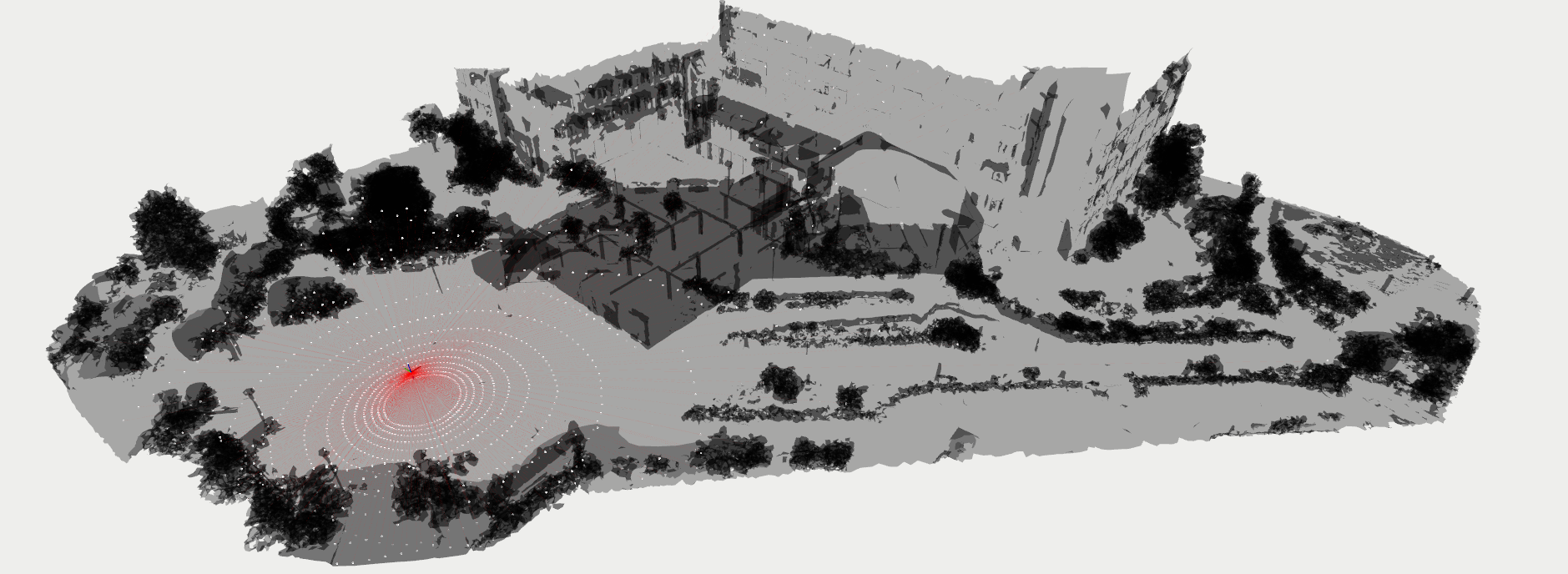} 
        \caption{Outdoor small}
        \label{fig:datasets:physics}
    \end{subfigure}
    \begin{subfigure}[t]{0.49\linewidth}
         \includegraphics[trim=0 0 0 0,clip,width=\linewidth]{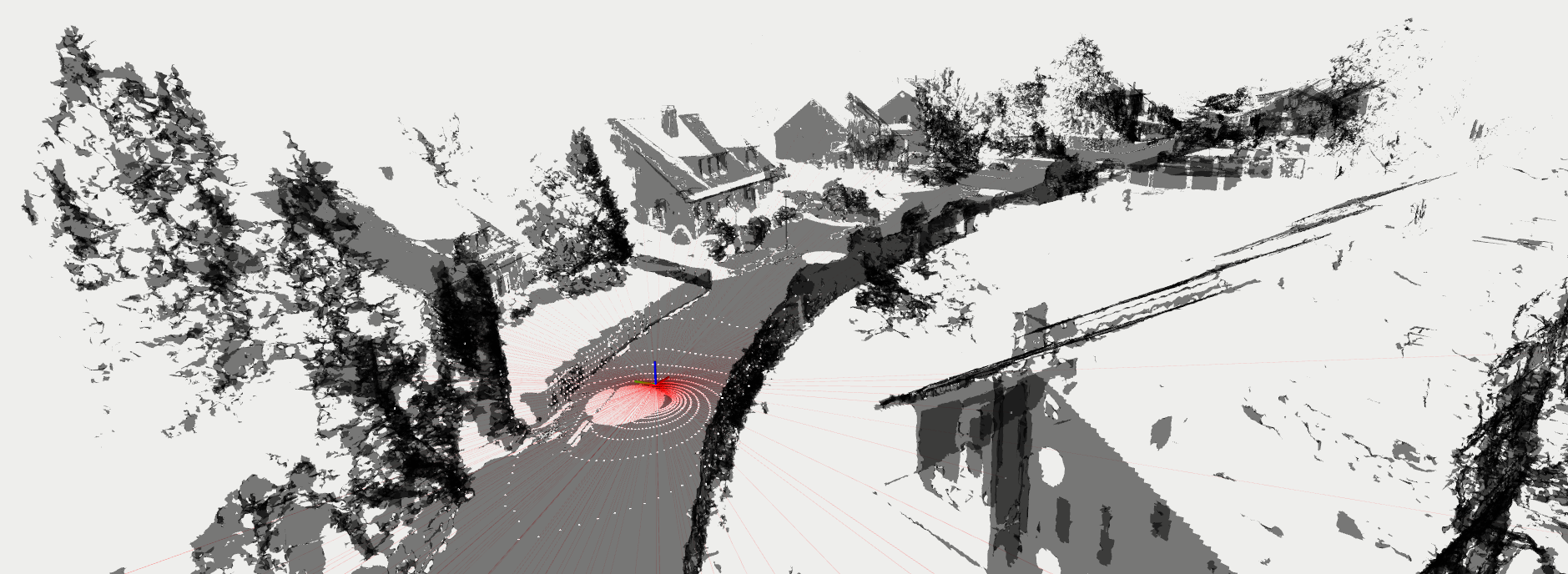}
         \caption{Outdoor large}
         \label{fig:datasets:westerberg}
    \end{subfigure}
    
    \caption{The real world maps used for evaluation. Example simulations of VLP-16 point clouds are rendered in red.}
    \label{fig:datasets}
    \vspace{-0.3cm}
\end{figure}

The first two experiments were done on synthetically generated maps.
In actual applications, maps of the real-world environments are more unstructured than the test data in the first experiments.
To account for that, we performed additional experiments on several real world maps of sites at Osnabrück University (indoor and outdoor), either manually built or reconstructed from point cloud data.
Example renderings of these data sets are shown in Fig.~\ref{fig:datasets}.
The left part of Tab.~\ref{tab:runtime_real} summarizes the size of the captured areas and the number of triangles in the maps.
The column "Data Type" specifies if the mesh was synthetically hand modeled or reconstructed with LVR2~\cite{wiemann2018irc}. 
Within each environment, we recorded the number of simulated VLP-16 scans that can be simulated in one second.
The results are summarized in Tab.~\ref{tab:runtime_real}.

\subsection{Quality}

\begin{table*}[t] 
    \centering
    \caption{Supported architectures, compute capabilities and memory location of simulation results.}
    \label{tab:compinst}
    \begin{tabular}{l c c c c c c c c c}
    \toprule
    & \multicolumn{3}{c}{Target Architecture} & \multicolumn{4}{c}{Compute Capabilities} & \multicolumn{2}{c}{Memory Location} \\
    \cmidrule(l{2pt}r{2pt}){2-4} \cmidrule(l{2pt}r{2pt}){5-8}\cmidrule(l{2pt}r{2pt}){9-10}
    & ARM\,x64 & CPU\,x86 & CPU\,x64 & CPU & GPU (-NV) & GPU (NV -RT) & GPU (NV RT) & CPU & GPU \\
    \cmidrule(l{2pt}r{2pt}){2-4} \cmidrule(l{2pt}r{2pt}){5-8}\cmidrule(l{2pt}r{2pt}){9-10}
    Gazebo    & \checkmark       & \checkmark & \checkmark & \checkmark & \checkmark & \checkmark &            & \checkmark & \\
    ISAAC Sim &            &             & \checkmark &            &            &            & \checkmark & \checkmark & \\
    Rmagine   & \checkmark & \checkmark & \checkmark  & \checkmark &            & \checkmark & \checkmark & \checkmark & \checkmark \\ 
    \bottomrule
    \end{tabular}
    \vspace{-0.3cm} 
\end{table*}

\begin{table*}[t] 
    \centering
    \caption{Evaluation in real world datasets for simulated Velodyne VLP-16 sensors in scans per second.}
    \label{tab:runtime_real}
    \vspace{-0.3cm} 
    \begin{tabular}{l r c r r r r r r r r r r} \\
    \toprule
    \multicolumn{1}{c}{} & \multicolumn{3}{c}{Dataset specs} & \multicolumn{5}{c}{CPU} & \multicolumn{4}{c}{GPU}\\
    \cmidrule(l{2pt}r{2pt}){2-4} \cmidrule(l{2pt}r{2pt}){5-9} \cmidrule(l{2pt}r{2pt}){10-13}
                                  & \multicolumn{1}{c}{Faces} & \multicolumn{1}{c}{Extension [m]} & \multicolumn{1}{c}{Data Type}
                                  & \multicolumn{1}{c}{NX}   & \multicolumn{1}{c}{X1}   & \multicolumn{1}{c}{NUC}  & \multicolumn{1}{c}{PC}    & \multicolumn{1}{c}{Legion} & \multicolumn{1}{c}{X1}    & \multicolumn{1}{c}{NUC}    & \multicolumn{1}{c}{PC}     & \multicolumn{1}{c}{Legion} \\
    \cmidrule(l{2pt}r{2pt}){2-4} \cmidrule(l{2pt}r{2pt}){5-9} \cmidrule(l{2pt}r{2pt}){10-13}
    \ref{fig:datasets:avz}   & 54,702 & $48\times28\times7$ & Synthetic     & 581 & 2,767 & 3,089 & 6,424 & 6,704  & 7,422 & 209,700 & 292,534 & 233,477 \\
    \ref{fig:datasets:berg}  & 353,189 & $22 \times 24 \times 6$ & Real     & 455  & 2,047 & 2,129 & 4,898 & 5,029  & 3,430  & 176,262 & 249,735 & 211,103 \\
    \ref{fig:datasets:physics} & 811,236 & $162 \times 84 \times 26$  & Real   & 390  & 1,774 & 1,832 & 4,119  & 4,100   & 2,260  & 151,885 & 214,348 & 191,800 \\
    \ref{fig:datasets:westerberg} & 19,716,244 & $279 \times 190 \times 41$ & Real & 197  & 1,186 & 964 & 2,029  & 2,148   & 1,512  &  44,577 & 84,193 & 114,112 \\
    \bottomrule
    \end{tabular}
    \vspace{-0.2cm}
\end{table*}

In the last experiment we compared the quality of the LiDAR scans simulated by Rmagine, Isaac Sim, and Gazebo.
To compare the simulations to a shared baseline, we also placed a real Velodyne VLP-16 in a scene that looks similar to the virtual scene and recorded its measurements.

\begin{figure}[H]
    \centering
    \begin{subfigure}[t]{0.325\linewidth}
        \includegraphics[width=\linewidth]{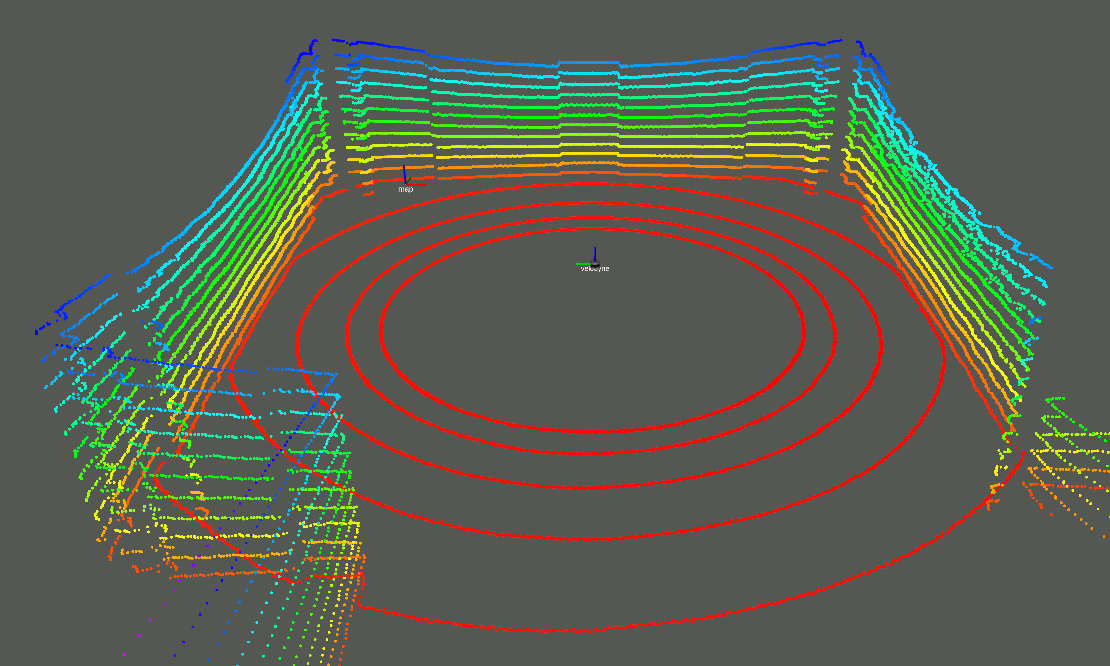}
        \caption{Real Scan}
        \label{fig:quality:real}
    \end{subfigure}
    \begin{subfigure}[t]{0.325\linewidth}
        \includegraphics[width=\linewidth]{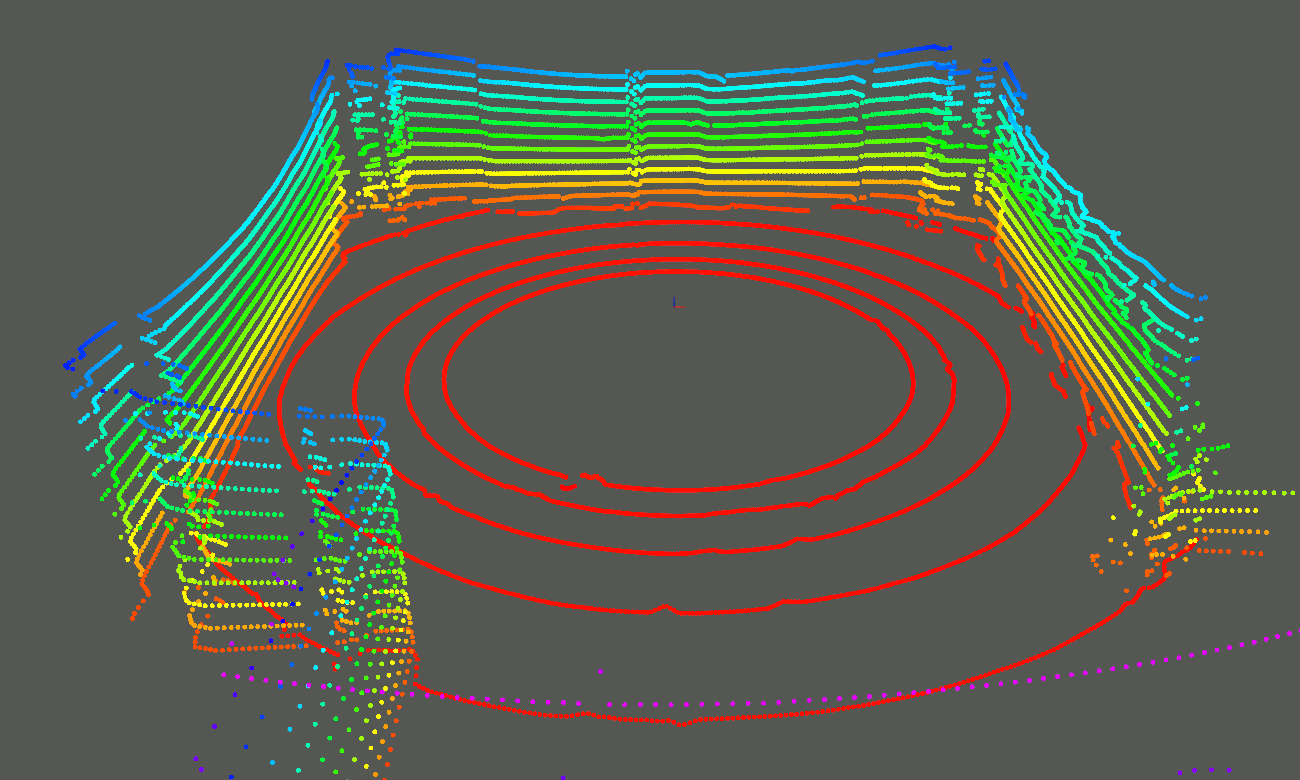} 
        \caption{Isaac Sim}
        \label{fig:quality:isaac}
    \end{subfigure}
    \begin{subfigure}[t]{0.325\linewidth}
        \includegraphics[width=\linewidth]{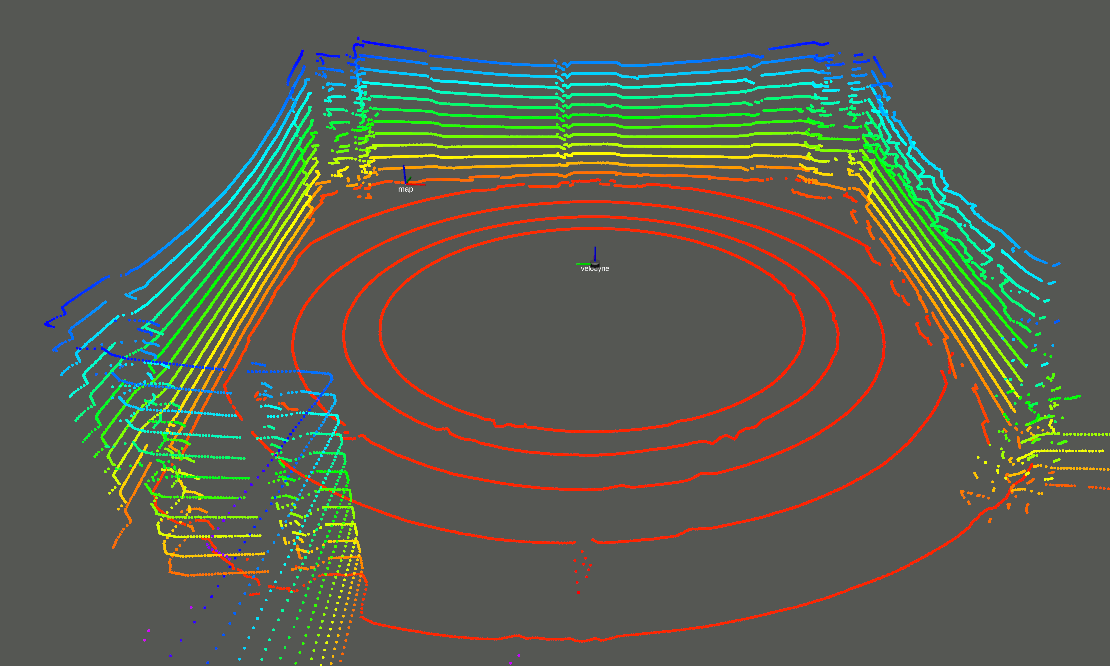}
        \caption{Gazebo CPU}
        \label{fig:quality:gzcpu}
    \end{subfigure} \\
    \begin{subfigure}[t]{0.325\linewidth}
        \includegraphics[width=\linewidth]{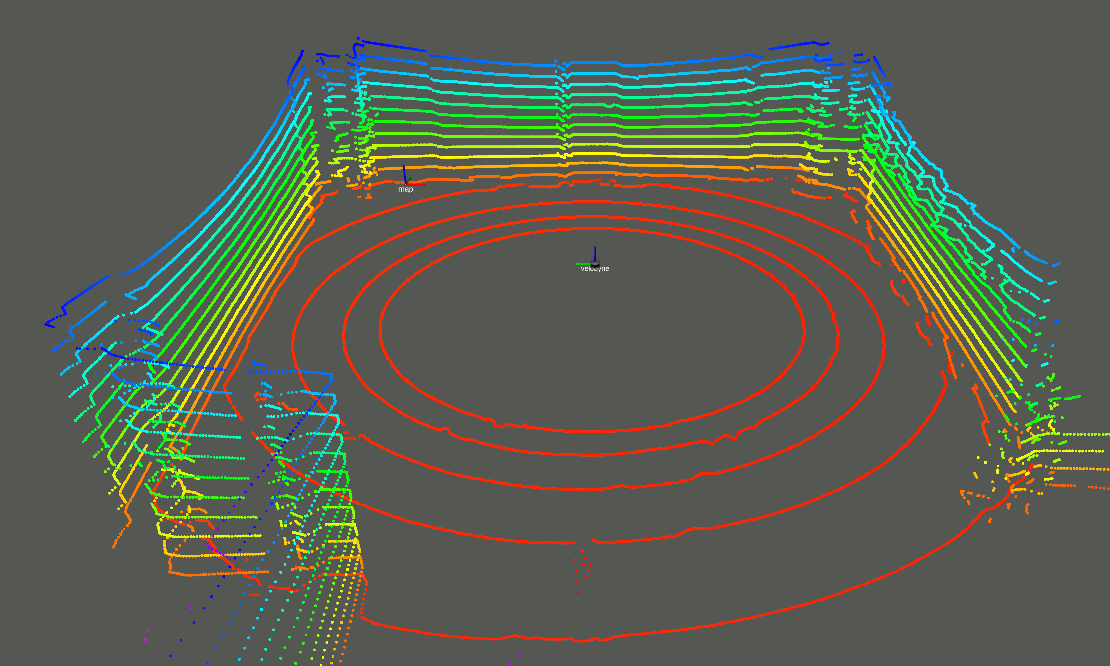}
        \caption{Gazebo GPU}
        \label{fig:quality:gzgpu}
    \end{subfigure} 
    \begin{subfigure}[t]{0.325\linewidth}
        \includegraphics[width=\linewidth]{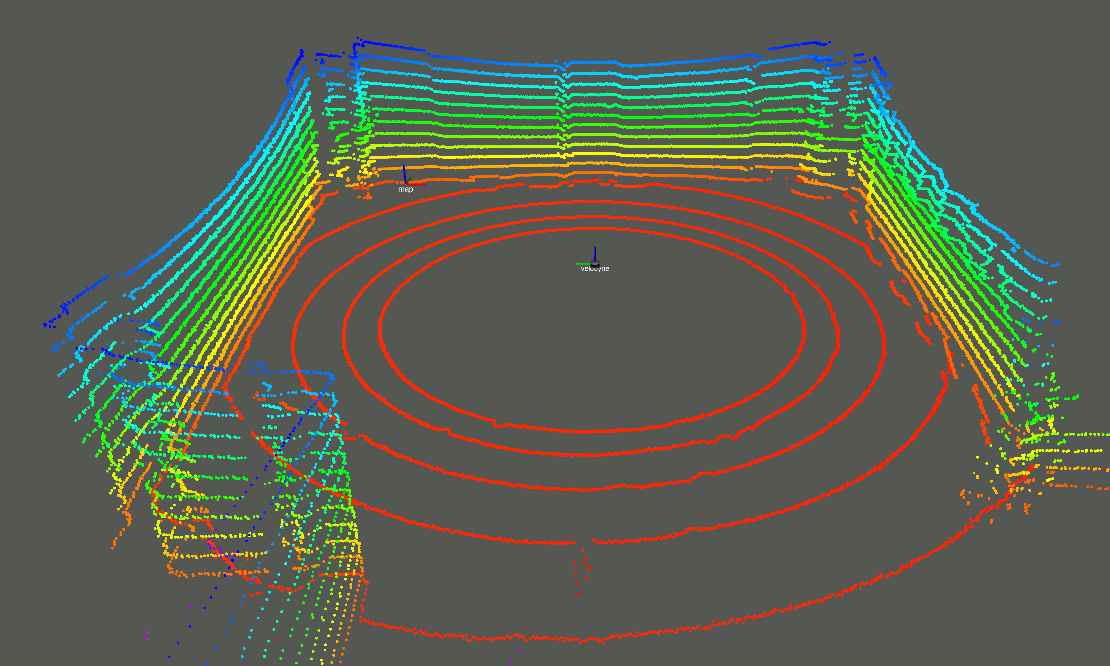} 
        \caption{Rmagine Embree}
        \label{fig:quality:rmembree}
    \end{subfigure}
    \begin{subfigure}[t]{0.325\linewidth}
         \includegraphics[width=\linewidth]{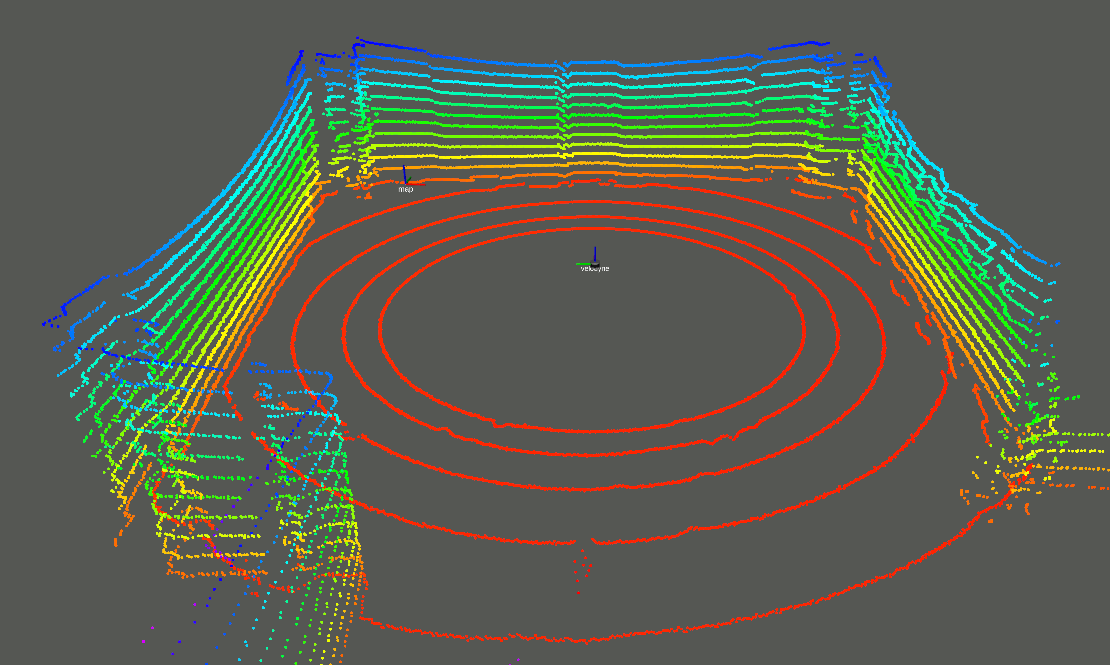}
         \caption{Rmagine OptiX}
         \label{fig:quality:rmoptix}
    \end{subfigure}
    
    \caption{Quality of LiDAR data simulated in map \ref{fig:datasets:berg} compared to real data of a Velodyne VLP-16 (a). There are no noticeable differences.}
    \label{fig:quality}
    \vspace{-0.5cm}
\end{figure}

\begin{figure}[H]
    \centering
    \includegraphics[width=\linewidth]{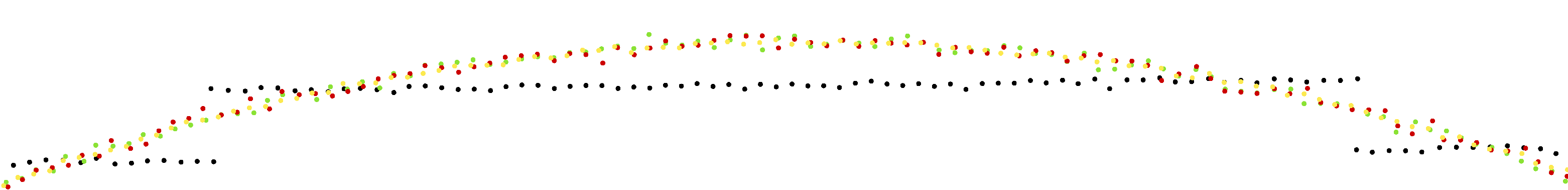}
    \caption{Comparison of Gazebo and Rmagine LiDAR simulations.
        Here, a Velodyne VLP-16 was placed 0.5 meters above a virtual ground plane.
        The image shows a part of the 6th scan line simulated by Gazebo on CPU (yellow), 
        Gazebo on GPU (black), Rmagine with Embree backend (red), and Rmagine with OptiX backend (green). }
    \label{fig:gzbug}
\end{figure}


For that, we chose the office environment of \ref{fig:datasets:berg}.
The results show that the quality of the Rmagine simulations such as Isaac Sim and Gazebo are similar to the real scan. 
The quality of simulation also depends heavily on the extent to which the underlying scene matches the real world.
Given the same scene of the environment, the comparison of the virtual scans shows a lot of similarities except for the GPU ray sensor of Gazebo (v11.9.1) as presented in Fig.~\ref{fig:gzbug}.
These anomalies only occur at rays hitting the surface at sharp angles.

\section{CONCLUSION AND FUTURE WORK}

We presented a library to simulate range sensor data on different platforms called Rmagine.
In contrast to established simulators, Rmagine is developed to directly run on the embedded computing devices of a mobile robot.
Rmagine makes the simulation of arbitrary range sensors feasible for future development of robotic algorithms. 
The evaluation demonstrated that our approach is fast compared to established simulators even without GPU acceleration and that it scales well to large maps.
The possibility to keep the simulation results in GPU memory allows to directly use the computed 3D data in GPU-accelerated algorithms.
Future work should enhance the library by additional map representations such as 3D occupancy grids, TSDF fields or 2D occupancy grids to be downwards compatible.
Furthermore, we aim to support GPUs of other manufacturers from AMD or Intel by implementing an OpenCL interface.
In principle, it might even be possible to integrate more application-specific devices like FPGAs.
We also aim to integrate materials and textures to allow developers model reflections and noise more accurately and to support colors, e.g, for algorithms that rely on image features.

\newpage








\bibliographystyle{IEEEtran}
\bibliography{IEEEabrv, references}

\balance

\end{document}